\documentclass[conference]{IEEEtran}
\usepackage[letterpaper, top=0.75in, bottom=1.1in, left=0.64in, right=0.64in]{geometry}
\IEEEoverridecommandlockouts
\usepackage{cite}
\usepackage{amsmath,amssymb,amsfonts}
\usepackage{graphicx}
\usepackage{amssymb}
\usepackage{textcomp}  
\usepackage{amsmath}
\usepackage{underscore}
\usepackage{xcolor}
\renewcommand{\figurename}{Figure}
\usepackage{algorithm}
\usepackage{array}
\usepackage[caption=false,font=normalsize,labelfont=sf,textfont=sf]{subfig}
\usepackage{textcomp}
\usepackage{stfloats}
\usepackage{url}
\usepackage{verbatim}
\usepackage{graphicx}
\usepackage{tabularx}
\usepackage{booktabs}
\usepackage{threeparttable}
\usepackage{balance}
\usepackage{multirow}
\usepackage{multicol}
\usepackage{enumitem}
\usepackage[skip=0pt]{caption}
\usepackage{subfig}      
\usepackage{algpseudocode}
\usepackage{epstopdf}
\usepackage{setspace}
\def\BibTeX{{\rm B\kern-.05em{\sc i\kern-.025em b}\kern-.08em
    T\kern-.1667em\lower.7ex\hbox{E}\kern-.125emX}}

\makeatletter
\newcommand{\linebreakand}{%
  \end{@IEEEauthorhalign}
  \hfill\mbox{}\par
  \mbox{}\hfill\begin{@IEEEauthorhalign}
}
\makeatother

\begin{document}

\title{Known Meets Unknown: Mitigating Overconfidence in Open Set Recognition}

\author{
Dongdong Zhao\textsuperscript{1}, 
Ranxin Fang\textsuperscript{1}, 
Changtian Song\textsuperscript{1}, 
Zhihui Liu\textsuperscript{2}, 
Jianwen Xiang\textsuperscript{1*}\thanks{*Corresponding authors} \\[1ex]
\textsuperscript{1}Wuhan University of Technology, China \\
\textsuperscript{2}Wuhan University, China \\[0.5ex]
\{zdd, star\_fang0916, songchangtian, zhihuil, jwxiang\}@whut.edu.cn\\[0.5ex]
}

\maketitle

\begin{abstract}
Open Set Recognition (OSR) requires models not only to accurately classify known classes but also to effectively reject unknown samples. However, when unknown samples are semantically similar to known classes, inter-class overlap in the feature space often causes models to assign unjustifiably high confidence to them, leading to misclassification as known classes---a phenomenon known as overconfidence. This overconfidence undermines OSR by blurring the decision boundary between known and unknown classes. To address this issue, we propose a framework that explicitly mitigates overconfidence caused by inter-class overlap. The framework consists of two components: a perturbation-based uncertainty estimation module, which applies controllable parameter perturbations to generate diverse predictions and quantify predictive uncertainty, and an unknown detection module with distinct learning-based classifiers, implemented as a two-stage procedure, which leverages the estimated uncertainty to improve discrimination between known and unknown classes, thereby enhancing OSR performance. Experimental results on three public datasets show that the proposed framework achieves superior performance over existing OSR methods.
\end{abstract}


\section{Introduction}
Traditional classification tasks typically focus on developing models that can accurately assign input samples to predefined classes \cite{neal2018open, mahdavi2021survey}. However, in real-world application scenarios, these models often encounter classes that were not present during training \cite{scheirer2012toward, oza2019c2ae, zhou2021learning}. This situation constitutes the so-called open set problem. Open Set Recognition (OSR) is a technique used in machine learning and pattern recognition to address this issue, aiming to endow models with the ability to detect and adapt to the emergence of new classes \cite{geng2021recent}. Specifically, the goal of OSR is to enable the model to not only accurately identify known classes but also effectively detect and reject samples of those unknown classes \cite{barcina2024managing}. 

Currently, various OSR methods \cite{zhang2020autonomous, yang2021deep, jang2022collective, liu2023learning} have been proposed for identifying unknown samples. Among these, the most popular method is to set a confidence threshold based on classifier outputs. The minimum confidence score produced by the classifier on known-class data is utilized as a classification threshold, and a test sample is subsequently classified as unknown if its confidence score is below this threshold \cite{zhang2020autonomous}. 

However, such an approach has a major limitation: due to the fact that models only focus on known classes during training, their predictions for unknown classes often lack mechanisms to regulate their confidence, when the features of unknown classes are similar to those of known classes, the model may produce high confidence scores for these samples, leading to misclassification in which unknown samples are incorrectly recognized as known classes \cite{nguyen2015deep, dhamija2018advances} and resulting in overconfidence \cite{neal2018open, geng2021recent}. To address overconfidence in OSR, several recent studies have attempted to mitigate it. For instance, EUE \cite{le2024deep} introduces evidence theory into the task of unknown recognition, under the intuition that known classes have sufficient evidence while unknown classes lack evidence. However, when unknown samples are semantically similar to certain known classes, inter-class overlap in the feature space can cause the model to collect seemingly sufficient evidence from these similar features. As a result, EUE may incorrectly classify such unknown samples as known, indicating that it cannot fully resolve overconfidence caused by inter-class overlap. 

Therefore, in this paper, we propose a framework for OSR that explicitly mitigates overconfidence caused by inter-class overlap. The framework consists of two main components. The first component aims to quantify predictive uncertainty, addressing the fact that models trained only on known classes lack mechanisms to regulate the confidence of unknown samples. To this end, controllable perturbations are applied to the trained model to generate diverse predictions. This step is motivated by the manifold theory in deep learning \cite{bengio2013representation}, which suggests that models capture the manifold structure of known classes by learning their data distribution \cite{huang2022class}. Applying perturbations can disrupt the learned manifold, causing predictable changes in outputs for known samples \cite{fawzi2018adversarial, bietti2019inductive, wang2022stability}, while the outputs for unknown samples exhibit unpredictable fluctuations with an average of zero \cite{li2023mope, mitchell2023detectgpt}. The variation in outputs under perturbation provides a principled signal for quantifying predictive uncertainty, capturing the reliability of model confidence for both known and unknown samples, which is subsequently leveraged in the second component of our framework to improve separability between known and unknown classes. In the second component, the estimated predictive uncertainty is leveraged in a two-stage unknown detection process. Specifically, the uncertainty serves as a signal to distinguish samples that are potentially unknown, and the two-stage design allows the model to first identify clear-cut unknowns and then refine decisions for ambiguous samples that lie near the decision boundary of known classes. This mechanism enhances the separability between known and unknown samples, thereby improving overall open set recognition performance. By integrating uncertainty estimation with this two-stage detection strategy, the framework effectively alleviates overconfidence caused by inter-class overlap. In summary, the main contributions of this paper are as follows:
\begin{itemize}
    \item We propose a novel OSR framework that explicitly alleviates overconfidence caused by inter-class overlap, thereby enhancing the capability of the model to identify unknown samples.
    \item To the best of our knowledge, this is the first work that explicitly leverages model perturbation to estimate predictive uncertainty in OSR, providing a principled way to quantify the reliability of model confidence.
    \item Extensive experiments on three public benchmarks demonstrate that the proposed framework outperforms existing state-of-the-art OSR methods.
\end{itemize}

\section{Related Work}
According to previous surveys on OSR \cite{mahdavi2021survey, vaze2022openset, barcina2024managing}, existing methods can be broadly categorized into generative and discriminative approaches.

Generative approaches aim to synthesize unknown class samples to enhance representation of unknown data during training, thereby improving recognition robustness in open environments. For instance, G-OpenMax \cite{ge2017generative} and OSRCI \cite{neal2018open} leveraged synthetic unknown samples to help the classifier learn a more distinct representation for separating known and unknown classes. However, these methods rely on complex training and assumptions about unknown distributions, and the generated samples may not accurately represent real unknown data. As a result, most recent OSR methods focus on discriminative approaches, which are also the focus of this paper.

Discriminative approaches distinguish known and unknown classes by learning the feature distribution of known data, using traditional machine learning or deep neural networks \cite{barcina2024managing}. Early work \cite{scheirer2012toward, scheirer2014probability} focused on extending traditional classifiers to handle unknown classes by introducing the concept of open space risk. Later, as research progressed, Extreme Value Theory (EVT) was introduced, and researchers \cite{rudd2017extreme, ping2022open} began to explore EVT-based frameworks to more effectively reject unknowns. However, these methods often perform poorly in high-dimensional feature spaces, limiting their scalability to complex data. Therefore, Deep Neural Networks (DNNs) have become the dominant choice for OSR, due to their ability to learn rich feature representations \cite{barcina2024managing}. Nevertheless, DNNs used for classification typically include a softmax layer, and studies \cite{bendale2016towards, wei2022mitigating} have shown that relying on softmax probabilities to identify unknown classes is unreliable. To address this issue, clustering-based methods and prototype-based methods have been proposed. Clustering-based methods \cite{henrydoss2020enhancing, sheng2023unknown} group samples of known classes and detect unknowns based on distances to cluster centers or cluster densities, while prototype-based methods \cite{yang2020convolutional, liu2023learning} learn representative prototypes for each known class and detect unknowns via distance or matching probability to these prototypes. Despite their effectiveness, both methods struggle when unknown samples overlap with known classes, limiting recognition accuracy. This limitation arises because these methods fundamentally rely on distances to cluster centers or prototypes, making them unable to reliably separate unknown samples that fall within or near dense regions of known-class features, which in turn can lead to overconfident misclassifications.

\section{Methodology}

\begin{figure*}[ht]
\centering
\includegraphics[width=1.0\textwidth, trim=10 60 46 60, clip]{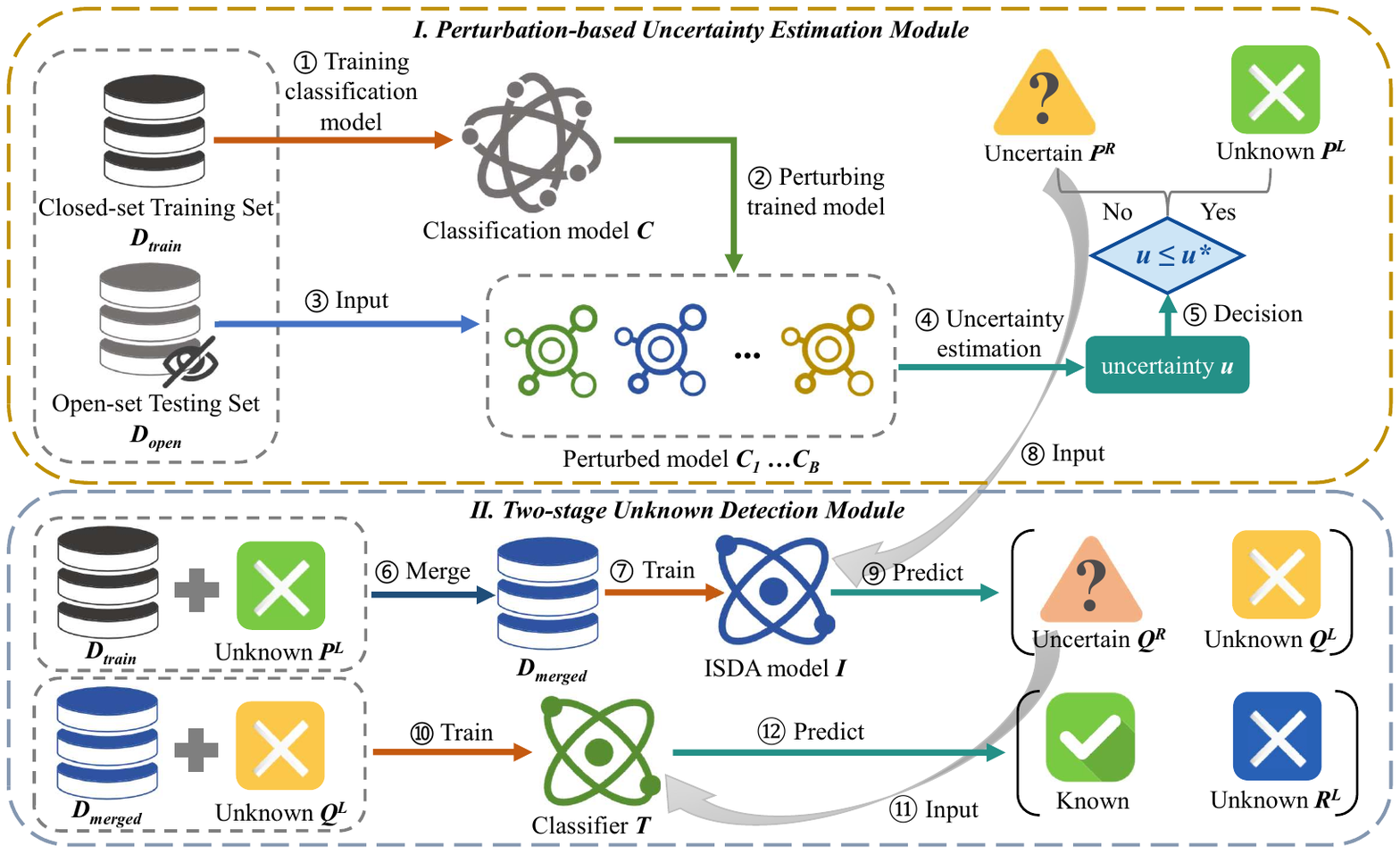}
\vspace{0.5em}
\caption{The overall architecture of the proposed framework. It consists of two main components: the perturbation-based uncertainty estimation module and the two-stage unknown detection module. The first component perturbs the trained classification model in the parameter space to estimate predictive uncertainty, while the second component leverages the estimated uncertainty to identify unknown samples through a two-stage decision process.}
\label{fig1}
\end{figure*}

\subsection{Problem Statement}
Before introducing the proposed framework, we briefly formalize the problem of OSR. Let $D_{train}=\{(x_i, y_i) \mid i = 1, \ldots, M\}$ be a set of $M$ labeled samples, where $x_i$ is the $i$-th training sample and $y_i \in \{1, \ldots K\}$ is the corresponding label. In the OSR phase, the data is unlabeled from the perspective of the model and drawn from a broader distribution. $D_{open}=\{(x_j, y_j) \mid j = 1, \ldots, N\}$ be a set of $N$ samples with ground-truth labels, where $y_j \in \{1, \ldots, K, K+1\}$. Here, labels $\{1, \ldots, K\}$ correspond to known classes (seen during training), while $y_j = K+1$ denotes the unknown class (not present in training). The objective of this paper is to accurately classify samples from known classes while identifying and rejecting those from unknown classes. The architecture of our proposed framework is illustrated in Figure \ref{fig1}, and the detailed descriptions of each component are provided in the following sections.

\subsection{Perturbation-based Uncertainty Estimation}
\begin{figure}[t]
    \centering
    \includegraphics[width=0.47\textwidth, trim=5 0 0 0, clip]{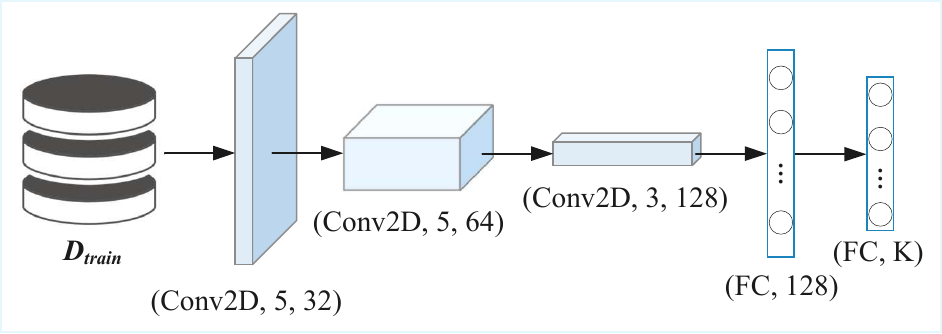}
    \vspace{1em}
    \caption{Classification model architecture.}
    \label{fig2}
\end{figure}

To satisfy the fundamental requirement of accurately recognizing known classes in OSR, we first construct a deep learning model based on Convolutional Neural Networks (CNNs), which serves as the foundation for subsequent stages. Figure \ref{fig2} details the model architecture, with the size of each layer annotated in parentheses below the corresponding layer in the diagram. The input layer is sequentially processed by multiple convolutional and activation layers to extract hierarchical feature representations. Each convolutional block consists of a convolutional operation followed by a ReLU activation, enabling nonlinear feature transformation. The architecture serves as a general backbone and can be adapted to different datasets. We emphasize that this architecture is highly common and has been widely adopted with minor differences such as hyperparameter selection and layer size in numerous studies \cite{wang2017malware, guo2017deep}. In addition, to optimize the model for the classification task, we minimize the cross-entropy loss, which measures the discrepancy between the predicted class probabilities and the ground truth labels.

After training, we obtain the classification model $C$, and then apply multiple independent perturbations of the same magnitude to its parameters, generating a set of predictions for each input. The variability among these predictions reflects the confidence of the model in its outputs, and we use predictive uncertainty to represent this confidence.

Specifically, we implement the perturbation mechanism by injecting Gaussian noise into the model parameters, which provides a simple yet effective way to simulate small and isotropic disturbances in the parameter space, consistent with the smooth manifold assumption in deep learning. By tuning the noise variance, we can flexibly adjust the perturbation strength to achieve a desirable trade-off between model stability and sensitivity.

The generated perturbations $\varepsilon$ follow a normal distribution:
\begin{equation}
    \varepsilon \sim \mathcal{N}(0, \sigma^2)
\end{equation}
where $\mathcal{N}(\cdot)$ represents a normal distribution, and $\sigma$ is the standard deviation used for generating Gaussian noise, with the specific formula given by:
\begin{equation}
    \sigma = \lambda\sqrt{\frac{1}{n}\sum_{i=1}^n(\theta_i-\overline{\theta})^2}
\end{equation}
where $\theta$ represents the parameters in a certain layer of the model, which are assumed to form a one-dimensional tensor (vector) of length $n$, $\theta_i$ denotes the $i$-th element of the parameter vector $\theta$, $\lambda$ is a hyperparameter used as a scaling factor to control the extent of noise, and $\overline{\theta}$ is the mean of the parameters $\theta_i$, which is defined as:
\begin{equation}
    \overline{\theta} = \frac{1}{n}\sum_{i=1}^n\theta_i
\end{equation}

For each layer, perturbations are applied separately using the same formulation. By controlling the number of perturbations $B$, we generate a set of perturbed models, denoted as $C_1, \ldots, C_B$. Each perturbed model processes the same input sample, and the resulting prediction ensemble is used to estimate the predictive uncertainty.

Specifically, we define the predictive uncertainty $\mu(x)$ as:
\begin{equation}
    \mu(x) = \left\| g\left( \frac{1}{B}\sum_{i=1}^B f_{\theta + \varepsilon_i}(x) \right) - g(f_\theta(x)) \right\|_2
\end{equation}
where $f_{\theta + \varepsilon_i}(x)$ is the output of the perturbed model $C_i$ for the sample $x$, while $f_\theta(x)$ is the output of the original model $C$, the function $g(\cdot)$ is defined as:
\begin{equation}
    g(p) = \log(p) - \log(1-p)
\end{equation}
where $p$ denotes a probability vector produced by a model. This transformation maps the softmax probabilities to logit space, linearizing the differences between probabilities under parameter perturbations, which allows the Euclidean norm to effectively measure the variation of the predictive distribution.

\begin{figure}[t]
\centering
\subfloat[Without inter-class overlap.]{
  \includegraphics[width=0.45\textwidth]{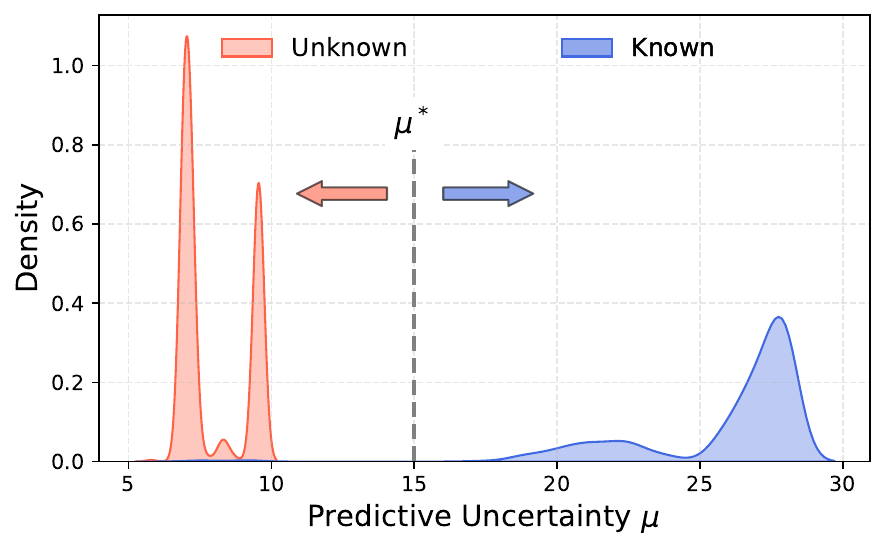}
  \label{fig:uncertainty_normal}
}
\vspace{0.4em}
\subfloat[With inter-class overlap.]{
  \includegraphics[width=0.45\textwidth]{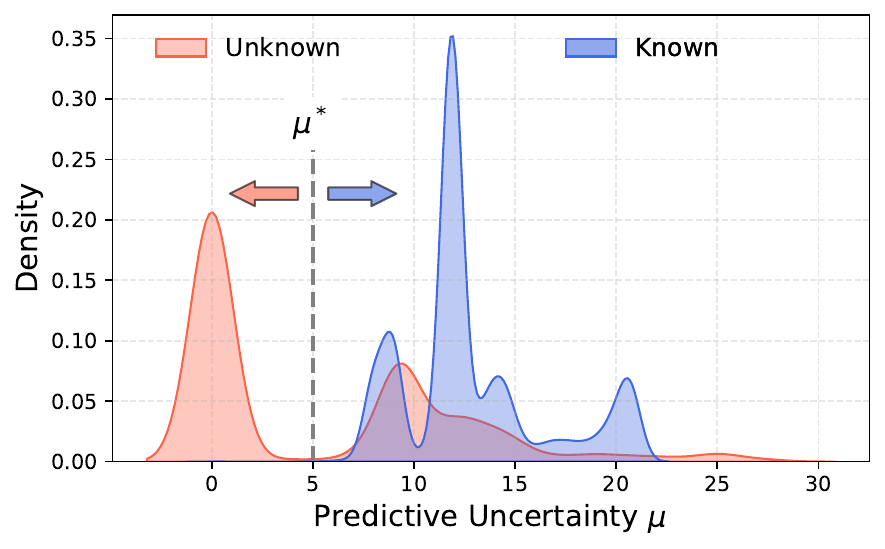}
  \label{fig:uncertainty_overlap}
}
\vspace{1em}
\caption{Density distributions of predictive uncertainty values for known and unknown samples under different data settings. }
\label{fig:fig3}
\end{figure}

To verify the effectiveness of the proposed predictive uncertainty measure, we analyze its distribution over known and unknown samples in real datasets. As illustrated in Figure \ref{fig:uncertainty_normal}, inputs from known classes exhibit larger variations under perturbations, resulting in higher predictive uncertainty, whereas inputs far from the known data distribution (unknown classes) produce more consistent predictions, leading to lower predictive uncertainty. This clear separation between known and unknown samples supports the effectiveness of the proposed predictive uncertainty measure and further supports the manifold-based motivation presented in the Introduction. 

In this way, we compute the predictive uncertainty value $\mu$ for each unlabeled sample and classify a sample as unknown if its uncertainty is less than or equal to a predefined threshold $\mu^\ast$.

\subsection{Two-stage Unknown Detection}
While the predictive uncertainty $\mu$ computed via parameter perturbations provides a principled measure for distinguishing many known and unknown samples, it may fail in cases where unknown inputs closely resemble known classes, as small variations introduced by perturbations may not produce sufficiently discriminative differences. This limitation stems from the inherent difficulty of overlapping class distributions in open set recognition, rather than a flaw in the uncertainty estimation itself. As shown in Figure \ref{fig:uncertainty_overlap}, when the dataset contains overlapping class regions, the distributions of predictive uncertainty for known and unknown samples exhibit noticeable overlap, reducing their separability. This observation further confirms that the degradation arises from the intrinsic entanglement of class manifolds. Under normal separability conditions, the uncertainty measure remains stable and effectively distinguishes known from unknown samples. Therefore, to address this limitation, we further introduce a two-stage unknown detection module, in which separate learning-based classifiers are trained at each stage to more effectively differentiate known from unknown samples.

\begin{algorithm}[t]
\caption{Two-stage Unknown Detection}
\label{alg:algorithm}
\begin{algorithmic}[1]
\Require Training set $D_{train}$, open set $D_{open}$, uncertainty values $\mu$, pretrained classifier $C$, threshold $\mu^\ast$
\Ensure Predicted labels

\State Initialize sets: $P^L, P^R, Q^L, Q^R, R^L \gets \emptyset$
\For{each sample $x_j \in D_{open}$}
    \If{$\mu_j \le \mu^\ast$}
        \State Assign label $K+1$ to $x_j$
        \State $P^L \gets P^L \cup x_j$
    \Else
        \State $P^R \gets P^R \cup x_j$
    \EndIf
\EndFor

\State Construct $D_{merged} = \{D_{train}, 0\} \cup \{P^L, 1\}$
\State Train ISDA $I(\cdot)$ by $D_{merged}$

\For{each sample $p_i \in P^R$}
    \If{$I(p_i) == 1$}
        \State Assign label $K+1$ to $p_i$
        \State $Q^L \gets Q^L \cup p_i$
    \Else
        \State $Q^R \gets Q^R \cup p_i$
    \EndIf
\EndFor

\State Update $D_{merged} \gets D_{merged} \cup \{Q^L, 1\}$
\State Train classifier $T(\cdot)$ by updated $D_{merged}$

\For{each sample $q_i \in Q^R$}
    \If{$T(q_i) == 1$}
        \State Assign label $K+1$ to $q_i$
        \State $R^L \gets R^L \cup q_i$
    \Else
        \State Assign predicted label from model $C$ to $q_i$
    \EndIf
\EndFor
\end{algorithmic}
\end{algorithm}

In the first stage, we adopt an improved subclass discriminant analysis (ISDA) \cite{xu2025cross} as the foundation and extend it to the open-set detection scenario. Specifically, ISDA performs subclass-level distribution alignment in the feature space to reduce inter-class overlap and enhance inter-class separability. It introduces two key parameters: $H_1$ (the number of unknown-class subclasses, following the setting in \cite{xu2025cross}) and $H_2$ (the number of known-class subclasses, treated as a tunable parameter). Different from the original ISDA, our implementation replaces the hard K-means clustering with a Gaussian Mixture Model (GMM) to achieve probabilistic soft assignment, which produces smoother and more continuous decision boundaries. After subclass modeling, a Gaussian Naïve Bayes (GaussianNB) classifier is trained on the aligned features to provide calibrated probabilistic predictions for the subsequent stage. It is worth noting that ISDA operates under a distribution modeling paradigm rather than an explicit loss-driven optimization. By estimating and aligning class-conditional feature statistics (means and covariances), it forms discriminative feature distributions without the need for a conventional loss function. In the second stage, we refine the separation between known and unknown samples using the probabilistic outputs from Stage 1. Specifically, a Decision Tree (DT) is trained due to its strong interpretability and ability to model non-linear decision boundaries in a low-dimensional probability space. Unlike deep neural models, DTs do not require large-scale training data or complex optimization, making them efficient and robust for this refinement task. This stage complements the distribution modeling in Stage 1 by explicitly learning the boundary conditions between known and unknown regions, thereby improving detection stability and reducing the misclassification of unknown samples.

The detailed implementation of the proposed two-stage unknown detection module is summarized in Algorithm~\ref{alg:algorithm}. Given the training set $D_{train}$, the open set $D_{open}$ with corresponding uncertainty values $\mu$, and a pretrained classification model $C$, the procedure operates as follows. First, all samples in $D_{open}$ are partitioned into an initial unknown subset $P^L$ and a candidate subset $P^R$ using the threshold $\mu^\ast$. The ISDA model $I(\cdot)$ is then trained on a merged dataset $D_{merged}$, which is composed of the training set $D_{train}$ (labeled as known, 0) and the initial unknown subset $P^L$ (labeled as unknown, 1). The trained ISDA model is subsequently used to refine the remaining candidate subset $P^R$, dividing it into a refined unknown subset $Q^L$ and a residual candidate subset $Q^R$. The merged dataset $D_{merged}$ is then updated by incorporating the refined unknown subset $Q^L$ (labeled as unknown, 1). Next, a DT classifier is subsequently trained on the updated $D_{merged}$ using a consistent binary labeling scheme (0 for known and 1 for unknown), and is applied to classify the remaining candidate samples $Q^R$. Samples predicted as unknown form the final set $R^L$, while the others are regarded as known samples and are assigned their class labels according to the predictions of model $C$. This progressive refinement process ensures that all unlabeled samples are consistently identified and assigned reliable class predictions.

\section{Experiment}

\subsection{Experimental Setup}
\textit{1) Datasets:} 
To validate the effectiveness of our proposed framework, we conduct comprehensive experiments on three public datasets: SCADA \cite{lemay2016providing}, GAS \cite{morris2014industrial} and ELECTRA \cite{gomez2019generation}. For each dataset, 50\% of the classes are randomly designated as unknown classes, while the remaining classes are treated as known classes. Among which, for the known classes, the samples are partitioned into training, validation, and test sets following a 3:1:1 ratio. For the unknown classes, the samples are split into validation and test sets with a 1:4 ratio \cite{le2024deep}.

\textit{2) Evaluation Metrics:} 
We evaluate the OSR performance using five widely adopted metrics: Accuracy, Precision, Recall, $F_1$-score and True Detection Rate (TDR). In particular, Accuracy refers to the overall correctness of assigning known samples to their classes and recognizing unknown samples as unknown, while TDR quantifies the proportion of correctly identified unknown samples among all true unknown samples, reflecting the capability of the model to detect unknowns.

\textit{3) Baselines:}
We compare the proposed framework with five state-of-the-art OSR baselines, including OpenMax \cite{bendale2016towards}, GradBP \cite{yang2021deep}, OVRN-CD \cite{jang2022collective}, MGPL \cite{liu2023learning}, and PostMax \cite{cruz2024operational}. For GradBP, two variants are implemented following the original paper: GradBP-1, which computes the gradient threshold as the maximum value of the gradient, and GradBP-2, which defines the threshold as the square root of the sum of squared gradients. To ensure fairness, we follow the original settings and use the official implementations or released code of the compared methods. All results are averaged over five independent runs.

\subsection{ Implementation Details}
The classification model within the proposed framework is implemented in PyTorch and trained on a single NVIDIA GeForce RTX 4090 GPU for 200 epochs using the Adam optimizer. The batch size is set to 256, and the learning rate is 0.0001. The key hyperparameters $B$, $\lambda$, $\mu^\ast$, and $H_2$ are selected individually for each dataset based on validation performance through a lightweight grid search. Specifically, for the SCADA dataset, we set $B=9$, $\lambda=4$, $\mu^\ast=13.0$, and $H_2=2$; for the GAS dataset, $B=7$, $\lambda=0.3$, $\mu^\ast=4.5$, and $H_2=1$; and for the ELECTRA dataset, $B=9$, $\lambda=2$, $\mu^\ast=9.9$, and $H_2=5$. All experiments are conducted under the same optimization settings to ensure fair comparison across datasets.

\subsection{Quantitative Results}
As shown in Table \ref{table1}, the proposed framework achieves the best performance across most evaluation metrics on all three datasets. Specifically, on the SCADA dataset, it attains the best results in Accuracy of 99.96\%, Precision of 99.99\%, $F_1$-score of 99.94\%, and TDR of 100\%, outperforming the runner-up method OVRN-CD by 0.07\% in Accuracy, 0.28\% in Precision, 0.12\% in $F_1$-score, and 0.02\% in TDR. On the GAS dataset, it outperforms all competitors with remarkable scores, improving Accuracy by 6.22\%, Precision by 3.94\%, $F_1$-score by 4.91\%, and TDR by 6.80\% over the second-best MGPL, while achieving a higher 2.67\% Recall than OVRN-CD. On the ELECTRA dataset, it achieves the highest Accuracy of 99.66\%, Precision of 100\%, $F_1$-score of 99.61\%, and TDR of 100\%, surpassing the runner-up by 0.04\%, 0.28\%, 0.05\%, and 0.21\%, respectively. These comprehensive results clearly validate the superior performance of the proposed OSR framework compared to other advanced methods.

\begin{table}[t]
\centering
\resizebox{\columnwidth}{!}{%
\begin{tabular}{lccccc}
\toprule
\multirow{2.5}{*}{\textbf{Method}} & \multicolumn{5}{c}{\textbf{SCADA}} \\ 
\cmidrule(lr){2-6}
 & \textbf{Accuracy} & \textbf{Precision} & \textbf{Recall} & \textbf{$F_1$-score} & \textbf{TDR}\\
\midrule
OpenMax & 72.27 & 79.34 & 55.73 & 63.79 & 79.74\\
GradBP-1 & 84.32 & 87.97 & 63.34 & 72.82 & 93.78\\
GradBP-2 & 84.71 & 99.70 & 50.99 & 67.19 & 99.93\\
OVRN-CD & \underline{99.89} & \underline{99.71} & \textbf{99.94} & \underline{99.82} & 99.87\\
MGPL & 59.29 & 52.87 & 72.36 & 55.31 & 53.39\\
PostMax & 78.82 & 99.60 & 31.96 & 39.01 & \underline{99.98}\\
Ours & \textbf{99.96} & \textbf{99.99} & \underline{99.89} & \textbf{99.94} & \textbf{100}\\
\midrule
\multirow{2.5}{*}{\textbf{Method}} & \multicolumn{5}{c}{\textbf{GAS}} \\ 
\cmidrule(lr){2-6}
 & \textbf{Accuracy} & \textbf{Precision} & \textbf{Recall} & \textbf{$F_1$-score} & \textbf{TDR}\\
\midrule
OpenMax & 74.50 & 77.43 & 86.46 & 81.43 & 53.70\\
GradBP-1 & 63.73 & 68.93 & 80.80 & 73.84 & 34.02\\
GradBP-2 & 69.39 & 69.98 & 93.01 & 79.64 & 28.29\\
OVRN-CD & 67.17 & 66.77 & \underline{96.35} & 78.86 & 16.39\\
MGPL & \underline{89.05} & \underline{89.99} & 93.13 & \underline{91.48} & \underline{81.94}\\
PostMax & 60.34 & 68.14 & 77.06 & 71.53 & 31.24\\
Ours & \textbf{95.27} & \textbf{93.93} & \textbf{99.02} & \textbf{96.39} & \textbf{88.74}\\
\midrule
\multirow{2.5}{*}{\textbf{Method}} & \multicolumn{5}{c}{\textbf{ELECTRA}} \\ 
\cmidrule(lr){2-6}
 & \textbf{Accuracy} & \textbf{Precision} & \textbf{Recall} & \textbf{$F_1$-score} & \textbf{TDR}\\
\midrule
OpenMax & 63.94 & 57.56 & 83.96 & 67.02 & 48.72\\
GradBP-1 & \underline{99.62} & \underline{99.72} & \textbf{99.41} & \underline{99.56} & \underline{99.79}\\
GradBP-2 & 99.60 & \underline{99.72} & \underline{99.36} & 99.54 & \underline{99.79}\\
OVRN-CD & 68.74 & 74.08 & 52.58 & 60.24 & 81.02\\
MGPL & 86.69 & 98.37 & 70.50 & 81.91 & 99.00\\
PostMax & 69.55 & 93.54 & 32.24 & 46.79 & 97.88\\
Ours & \textbf{99.66} & \textbf{100} & 99.22 & \textbf{99.61} & \textbf{100}\\
\bottomrule
\end{tabular}%
}
\vspace{1em}
\caption{Performance comparison of methods on three datasets. The bold and underlined values indicate the best and runner-up results, respectively.}
\label{table1}
\end{table}

\subsection{Ablation Study}
We conduct an ablation study to evaluate the improvement in performance attributed to the two introduced stages. Specifically, we remove the first-stage ISDA classification and the second-stage DT classification separately. The results are reported in Table \ref{table2}. Here, ``Perturbation Only'' refers to using perturbation-based uncertainty to distinguish known and unknown samples, without employing the two-stage framework. We observe that applying either the first-stage ISDA classification, the second-stage DT classification, or both leads to consistent performance gains. These results demonstrate that integrating any part of the proposed framework contributes to improved overall accuracy. Especially, on the GAS dataset, the standalone use of either the ISDA stage or the DT stage leads to noticeable accuracy improvements, while the combined application of both stages achieves the highest accuracy, thereby demonstrating the effectiveness of the introduced two-stage on this dataset.

\begin{table}[t]
\centering
\begin{small}
\begin{tabular}{lccc}
\toprule
\textbf{Setting} & \textbf{SCADA} & \textbf{GAS} & \textbf{ELECTRA} \\
\midrule
Perturbation Only & 71.96 & 72.96 & 99.65\\
w/o ISDA & 99.96 & 82.42 & 99.67\\
w/o DT & 78.85 & 93.43 & 99.66\\
Ours & 99.96 & 95.27 & 99.66\\
\bottomrule
\end{tabular}
\end{small}
\vspace{1em}
\caption{Accuracy (\%) of the ablation study on three datasets.}
\label{table2}
\end{table}

\subsection{Parameter Sensitivity Analysis}
\begin{figure}[t]
    \centering
    \subfloat[Effects of $B$]{%
        \includegraphics[width=0.48\linewidth]{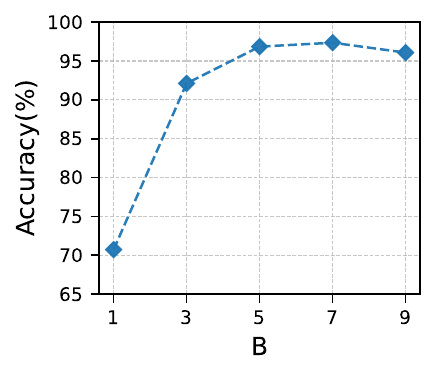}%
        \label{fig:sub1}
    }\hfill
    \subfloat[Effects of $\lambda$]{%
        \includegraphics[width=0.48\linewidth]{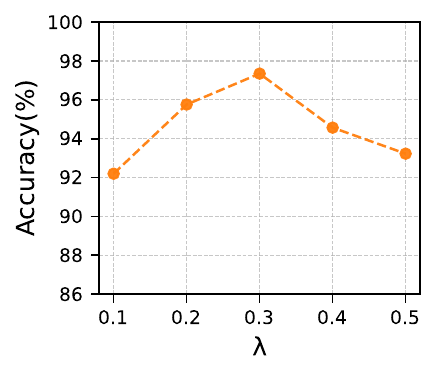}%
        \label{fig:sub2}
    }\hfill
    \subfloat[Effects of $\mu^{*}$]{%
        \includegraphics[width=0.48\linewidth]{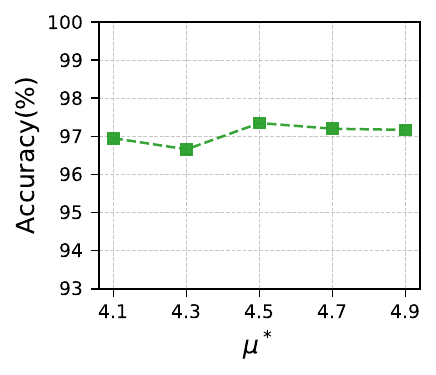}%
        \label{fig:sub3}
    }\hfill
    \subfloat[Effects of $H_2$]{%
        \includegraphics[width=0.48\linewidth]{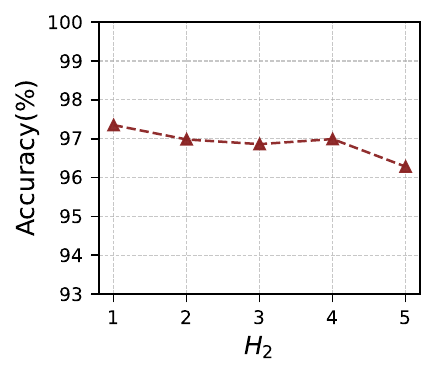}%
        \label{fig:sub4}
    }
    \vspace{1em}
    \caption{Parameter sensitivity analysis of $B$, $\lambda$, $\mu^{*}$, and $H_2$ on the GAS dataset.}
    \label{fig:figure4}
\end{figure}

We conduct parameter sensitivity analysis on four key hyperparameters, including perturbed model number $B$, noise scale factor $\lambda$, classification threshold $\mu^{\ast}$, and the number of known-class subclasses $H_2$. The analysis is conducted on the validation set, with Accuracy used as the evaluation metric. Figure \ref{fig:figure4} illustrates the results of the four hyperparameters on the GAS dataset.

\textit{1) Perturbed model number $B$:}
We first examine the effect of $B$ on OSR performance. Specifically, we vary $B$ from $\{1,3,5,7,9\}$ while keeping other hyperparameters fixed. As shown in Figure \ref{fig:sub1}, the accuracy increases with larger $B$, since multiple perturbations generate more diverse predictions and lead to a more reliable estimation of predictive uncertainty. However, when $B$ exceeds a certain threshold (e.g., $B=7$), the performance begins to decline because excessive perturbations introduce redundant information. Such redundancy not only fails to provide additional insight but also leads to over-smoothed uncertainty estimates, thereby reducing the distinction between known and unknown samples.

\textit{2) Noise scale factor $\lambda$:}
We next analyze the impact of the noise scale factor $\lambda$, which controls the magnitude of each perturbation. As shown in Figure \ref{fig:sub2}, increasing $\lambda$ initially improves accuracy, as moderate perturbations induce controlled changes in the outputs of known samples, which enhances the discrimination between known and unknown samples. However, when $\lambda$ becomes excessively large, the robustness of the model is completely compromised. It generates unstable outputs for known samples, leading to poor discrimination between known and unknown data and consequently degrading OSR performance.

\begin{figure}[t]
    \centering
    \subfloat[Effects of $H_2$]{%
        \includegraphics[width=0.48\linewidth]{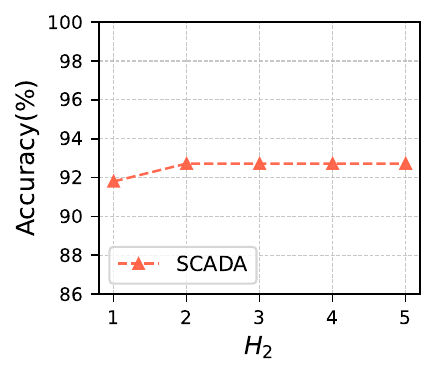}%
        \label{fig:sub5}
    }\hfill
    \subfloat[Effects of $H_2$]{%
        \includegraphics[width=0.48\linewidth]{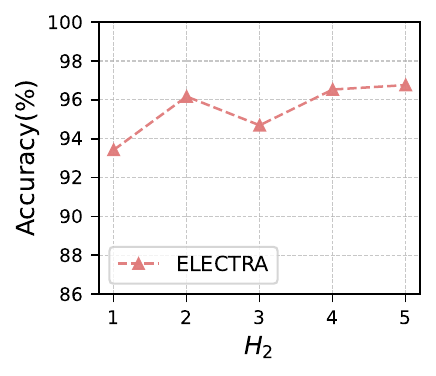}%
        \label{fig:sub6}
    }
    \vspace{1em}
    \caption{Accuracy variation with $H_2$ on SCADA and ELECTRA datasets.}
    \label{fig:figure5}
\end{figure}

\textit{3) Classification threshold $\mu^{\ast}$:}
The threshold $\mu^{\ast}$ is used to select high-quality unknown samples from incoming unlabeled data, which are guaranteed to be free of inter-class overlap with known classes in the feature space. Figure \ref{fig:sub3} illustrates examples of setting $\mu^{\ast}$ to 4.1, 4.3, 4.5, 4.7, and 4.9, respectively. The results indicate that the OSR capability of the proposed framework is relatively robust to variations in $\mu^{\ast}$, and the framework can reliably distinguish unknown samples without precise tuning of this threshold.

\textit{4) Number of known-class subclasses $H_2$:}
The hyperparameter $H_2$ controls the number of prototypes assigned to each known class in the ISDA feature space, determining how many subclasses each known class is split into. Figure \ref{fig:sub4} and Figure \ref{fig:figure5} show the performance trends for $H_2$ across three datasets, respectively. On the GAS dataset, the overall performance shows a decreasing trend as $H_2$ increases, indicating that the known classes are already compact and subdividing them fragments inter-class representation. In contrast, on the ELECTRA and SCADA datasets, increasing $H_2$ maintains or slightly improves performance, suggesting that the subdivision mechanism can be beneficial for datasets with higher inter-class variability. These results demonstrate that $H_2$ provides a flexible mechanism to adapt to different dataset characteristics.

\section{CONCLUSION}
In this paper, we propose a novel OSR framework that explicitly addresses the overconfidence problem caused by inter-class overlap. The framework consists of two key components: a perturbation-based uncertainty estimation module and a two-stage unknown detection module. The first module applies controlled perturbations to the trained classification model to quantify predictive uncertainty, allowing the identification of high-quality unknown samples that are clearly separable from known classes. These reliable unknown samples are then leveraged in a two-stage detection process to refine decision boundaries, effectively reducing the misclassification of unknown instances as known, which directly alleviates the overconfidence. Extensive experiments on three public benchmarks demonstrate that the proposed framework achieves superior performance compared with state-of-the-art OSR methods. While the perturbation-based uncertainty estimation involves multiple forward passes through the trained model, the approach remains practical for standard-scale applications. Future work will focus on exploring adaptive perturbation strategies and clustering-based techniques to further improve scalability and enable fine-grained recognition of unknown classes.

\end{document}